\documentclass[11pt]{article}
\usepackage{acl2016}
\usepackage{times}
\usepackage{url}
\usepackage{latexsym}
\usepackage{graphicx}
\usepackage{multirow}
\usepackage{fancyhdr}

\aclfinalcopy 

\title{Event Nugget Detection with Forward-Backward Recurrent Neural Networks\thanks{\ \ Published as a short paper at ACL 2016}}

\author{Reza Ghaeini, Xiaoli Z. Fern, Liang Huang, Prasad Tadepalli\\
School of Electrical Engineering and Computer Science, Oregon State University\\
1148 Kelley Engineering Center, Corvallis, OR 97331-5501, USA\\
{\tt \{ghaeinim, xfern, huanlian, tadepall\}@eecs.oregonstate.edu}}

\date{}

\begin{document}
\maketitle
\begin{abstract}
Traditional event detection methods heavily rely on manually engineered rich features. Recent deep learning approaches alleviate this problem by automatic feature engineering. But such efforts, like tradition methods, have so far only focused on single-token event mentions, whereas in practice events can also be a phrase. We instead use forward-backward recurrent neural networks (FBRNNs) to detect events that can be either words or phrases. To the best our knowledge, this is one of the first efforts to handle multi-word events and also the first attempt to use RNNs for event detection. Experimental results demonstrate that FBRNN is competitive with the state-of-the-art methods on the ACE 2005 and the Rich ERE 2015 event detection tasks.
\end{abstract}

\section{Introduction}

\begin{figure*}[t!]
\centering
\includegraphics[width=0.86\textwidth ]{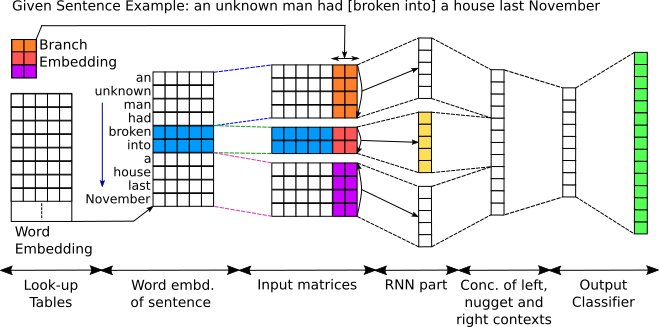}
\caption{The Proposed Forward-Backward
Recurrent Neural Network (FBRNN) Model, with the example sentence ``an unknown man had [broken into] a house last November'' and event nugget candidate ``broken into''}
\end{figure*}
Automatic event extraction from natural text is an important and challenging task for natural language understanding. Given a set of ontologized event types, the goal of event extraction is to identify the mentions of different event types and their arguments from natural texts. In this paper we focus on the problem of extracting event mentions, which can be in the form of a single word or multiple words. In the current literature, events have been annotated in two different forms: 
\vspace{-0.05in}
\begin{itemize}
\item \textbf{Event trigger}: a single token that is considered to signify the occurrence of an event. Here a token is not necessarily a word, for example, in order to capture a death event, the phrase ``kick the bucket'' is concatenated into a single token ``kick\_the\_bucket''. This scheme has been used in the ACE and Light ERE data and has been followed in most studies on event extraction.
\vspace{-0.05in}
\item \textbf{Event nugget}: a word or a phrase of multiple words that most clearly expresses the occurrence of an event. This scheme is recently introduced to remove the limitation of single-token event triggers and has been adopted by the rich ERE data for event annotation.
\end{itemize}
Existing event extraction work often heavily relies on a rich set of hand-designed features and utilizes existing NLP toolkits and resources~\cite{Ji_Grashman_2008,Patwardhan,Liao_2010,McClosky,Huang,Li_2013_a,Li_2013,Li_2014}.
Consequently, it is often challenging to adapt prior methods to multi-lingual or non-English settings since they require extensive linguistic knowledge for feature engineering and mature NLP toolkits for extracting the features without severe error propagation. 

By contrast, deep learning has recently emerged as a compelling solution to avoid the aforementioned problems by automatically extracting meaningful features from raw text without relying on existing NLP toolkits. There have been some limited attempts in using deep learning for event detection~\cite{CNN,DCNN} which apply Convolutional Neural Networks (CNNs) to a window of text around potential triggers to identify events. 
These efforts outperform traditional methods, but there remain two major limitations:
\vspace{-0.05in}
\begin{itemize}
\item So far they have, like traditional methods, only focused on the oversimplified scenario of single-token event detection. 
\vspace{-0.05in}
\item Such CNN-based approaches require a fixed size window. In practice it is often unclear how large this window needs to be in order to capture necessary context to make decision for an event candidate. 
\end{itemize}

Recurrent Neural Networks (RNNs), by contrast, is a natural solution to both problems above because it can be applied to inputs of variable length which eliminates both the requirement of single-token event trigger and the need for a fixed window size. Using recurrent nodes with Long Short Term Memory (LSTM)~\cite{LSTM} or Gated Recurrent Units (GRU)~\cite{Cho}, RNN is potentially capable of selectively deciding the relevant context to consider for detecting events.

In this paper we present a forward-backward recurrent neural network (FBRNN) to extract (possibly multi-word) event mentions from raw text. Although RNNs have been studied extensively in other NLP tasks~\cite{B_LSTM,Tai,Socher_2014,Paulus}, to the best of our knowledge, this is the first work to use RNNs for event detection. This is also one of the first efforts to handle multi-word event nuggets. Experimental results confirm that FBRNN is competitive compared to the state-of-the-art on the ACE 2005 dataset and the Rich ERE 2015 event detection task.

\section{Proposed Model}

Let $x = [w_{0}, w_{1}, ..., w_{n}]$ be a sentence. We first go over each word and phrase and heuristically extract a set of event candidates. The task is then to predict for each candidate given the sentence whether it is an event and, if so, its type. Figure 1 demonstrates our proposed model for this task.  

For each event candidate, which consists of a continuous span of texts $[w_{i}, ..., w_{j}]$, we split the sentence into three parts: the left context $[w_{0}, ..., w_{i-1}]$, the event nugget candidate $[w_{i}, ..., w_{j}]$ and the right context $[w_{j+1}, ..., w_{n}]$. For instance, for event candidate ``broken into'' and given sentence ``an unknown man had broken into a house last November''; [an, unknown, man, had], [broken, into] and [a, house, last, November] are the left context, the event nugget candidate and the right context respectively. For each part, we learn a separate RNN to produce a representation. Before feeding the data into the network, each word is represented as a real-valued vector that is formed by concatenating a word embedding with a branch embedding, which we describe below:
\begin{itemize}
\item \textbf{Word embedding}: Several studies have investigated methods for representing words as real-valued vectors in order to capture the hidden semantic and syntactic properties of words~\cite{Collobert_2008,word_2_vec}. Such embeddings are typically learned from large unlabeled text corpora, consequently can serve as good initializations. In our work, we initialize the word embedding with the pretrained 300-diemension word2vec~\cite{word_2_vec}.
\item \textbf{Branch embedding}: The relative position of a word to the current event nugget candidate may contain useful information toward how the word should be used or interpreted in identifying events. It is thus a common practice to include an additional embedding for each word that characterizes its relative position to the event nugget candidate. In this work, to reduce the complexity of our model and avoid overfitting, we only learn embeddings for three different positions: the left branch, the nugget branch and the right branch respectively. This is illustrated using three different colors in Figure 1.
\end{itemize}

Now each word is represented as a real-valued vector, formed by concatenating its word and branch embeddings. The sequence of words in the left, nugget and right branches will each pass through a separate Recurrent Neural Network. For the left and nugget branches, we process the words from left to right, and use the opposite direction (from right to left) for the right context, thus the name Forward-Backward RNN (FBRNN).

The output of each recurrent neural network is a fixed size representation of its input.  We concatenate the representations from the three branches and pass it through a fully connected neural network with a softmax output node that classifies each event candidate as an event of specific type or a non-event. Note that in cases where an event candidate can potentially belong to multiple event types, one can replace the softmax output node with a set of binary output nodes or a sigmoid to allow for multi-label prediction for each event candidate. 

To avoid overfitting, we use dropout~\cite{dropout_2,dropout} with rate of 0.5 for regularization. The weights of the recurrent neural networks as well as the fully connected neural network are learned by minimizing the log-loss on the training data via the Adam optimizer~\cite{ADAM} which performs better that other optimization methods like AdaDelta~\cite{AdaDelta}, AdaGrad~\cite{AdaGrad}, RMSprop and SGD. During training, the word and branch embeddings are updated to learn effective representations for this specific task.

\section{Experiments}
In this section, we first empirically examine some design choices for our model and then compare the proposed model to the current state-of-the-art on two different event detection datasets.

\subsection{Datasets, candidate generation and hyper-parameters}
We experiment on two different corpora, ACE 2005 and Rich ERE 2015. 
\begin{itemize}
\item \textbf{ACE 2005}: The ACE 2005 corpus is annotated with single-token event triggers and has eight event types and 33 event subtypes that, along with the ``non-event" class, constitutes a 34-class classification problem. In our experiments we used the same train, development and test sets as the previous studies on this dataset~\cite{CNN,Li_2013}. Candidate generation for this corpus is based on a list of candidate event trigger words created from the training data and the PPDB paraphrase database. Given a sentence, we go over each token and extract the tokens that appear in this high-recall list as event candidates, which we then classify with our proposed FBRNN model. 

\item \textbf{Rich ERE 2015}: The Rich ERE 2015 corpus was released in the TAC 2015 competition and annotated at the nugget level, thus addressing phrasal event mentions. The Rich ERE 2015 corpus has nine event types and 38 event subtypes, forming a 39-class classification problem (considering ``non-event'' as an additional class). We utilized the same train and test sets that have been used in the TAC 2015 event nugget detection competition. A subset of the provided train set was set aside as our development set. 
To generate event nugget candidates, we first followed the same strategy that we used for the ACE 2005 dataset experiment to identify single-token event candidates. We then expand the single-token event candidates using a heuristic rule based on POS tags. 

\end{itemize}

There are a number of hyper-parameters for our model, including the dimension of the branch embedding, the number of recurrent layers in each RNN, the size of the RNN outputs, the dropout rates for training the networks. We tune these parameters using the development set. 

\subsection{Exploration of different design choices}

\begin{table}
\small
\begin{center}
\begin{tabular}{|c|c|c|c|c|}
\hline 
\multicolumn{2}{|c|}{Configurations} & P & R & F1 \\ \hline 
\multirow{2}{*}{LSTM} & +branch & 59.82 & 48.39 & 53.50 \\ \cline{2-5}
  & -branch & 58.50 & 44.82 & 50.76 \\ \hline
\multirow{2}{*}{GRU} & +branch & 63.72 & 47.68 & \textbf{54.55}  \\ \cline{2-5}
 & -branch & 64.56 & 43.93 & 52.28 \\ \hline
\end{tabular}
\caption{Performance on the development set with different configurations on Rich ERE 2015.\label{tab:conf}}
\end{center}
\end{table}

We first design some experiments to evaluate the impact of the following design choices:
\begin{itemize}
\item[i)] Different RNN structures: LSTM and GRU are two popular recurrent network structures that are capable of extracting long-term dependencies in different ways. Here we compare their performance for event detection. 

\item[ii)] The effect of branch embedding: A word can present different role and concept when it is in a nugget branch or other branches. Here we would examine the effect of including branch embedding. 

\end{itemize}
Table~\ref{tab:conf} shows the results of our model with different design choices on the development set of the Rich ERE 2015 corpus. 
We note that the performance of GRU is slightly better than that of LSTM. We believe this is because GRU is a less complex structure compared to LSTM, thus less prone to overfitting given the limited training data for our task. From the results we can also see that the branch embedding performs a crucial role for our model, producing significant improvement for both LSTM and GRU. 
Based on the results presented above, for the remaining experiments we will focus on GRU structure with branch embeddings. 

\subsection{Results on ACE 2005}

\begin{table}
\small
\begin{center}
\begin{tabular}{|c|c|c|c|}
\hline
Methods & P & R & F1 \\ \hline
Sentence level in Ji and & \multirow{2}{*}{-} & \multirow{2}{*}{-} & \multirow{2}{*}{59.7} \\
Grishman (2008) & & & \\ \hline
MaxEnt with local & \multirow{2}{*}{-} & \multirow{2}{*}{-} & \multirow{2}{*}{64.7} \\ 
features in Li et al. (2013b) & & & \\ \hline
Joint beam search with local & \multirow{2}{*}{-} & \multirow{2}{*}{-} & \multirow{2}{*}{63.7} \\
features in Li et al. (2013b) & & & \\ \hline
Joint beam search with & \multirow{3}{*}{-} & \multirow{3}{*}{-} & \multirow{3}{*}{65.6} \\
local and global features in & & & \\ 
Li et al. (2013b) & & & \\ \hline\hline
CNN (Nguyen, 2015) & 71.9 & 63.8 & 67.6 \\ \hline\hline
FBRNN & 66.8 & 68.0 & 67.4 \\ \hline
\end{tabular}
\caption{Comparison with reported performance by event detection systems without using gold entity mentions and types on the ACE 2005 corpus.\label{tab:ace}}
\end{center}
\end{table}

Many prior studies employ gold-standard entity mentions and types from manual annotation, which would not be available in reality during testing. Nguyen and Grishman (2015) examined the performance of a number of traditional systems~\cite{Li_2013} in a more realistic setting, where entity mentions and types are acquired from an automatic high-performing name tagger and information extraction system. In Table~\ref{tab:ace} we compare the performance of our system with these results reported by Nguyen and Grishman (2015). 

We first note that the deep learning methods (CNN and FBRNN) achieve significantly better F1 performance compared to traditional methods using manually engineered features (both local and global). Compared to CNN, our FBRNN model achieved better recall but the precision is lower. For the overall F1 measure, our model is comparable with the CNN model.

\subsection{Results on Rich ERE 2015}

\begin{table}
\small
\begin{center}
\begin{tabular}{|c|c|c|c|}
\hline
Methods & P & R & F1 \\ \hline
$1^{st}$ & 75.23 & 47.74 & 58.41 \\ \hline
$2^{nd}$ & 73.95 & 46.61 & 57.18 \\ \hline
$3^{th}$ & 73.68 & 44.94 & 55.83 \\ \hline
$4^{th}$ & 73.73 & 44.57 & 55.56 \\ \hline
$5^{th}$ & 71.06 & 43.50 & 53.97 \\ \hline\hline
FBRNN & 71.58 & 48.19 & 57.61 \\ \hline
\end{tabular}
\caption{Performance of FBRNN compared with reported top results in TAC competition \cite{TAC_2015} on Rich ERE 2015.\label{tab:ere}}
\end{center}
\end{table}

Table~\ref{tab:ere} reports the test performance of our model and shows that it is competitive with the top-ranked results obtained in the TAC 2015 event nugget detection competition. It is interesting to note that FBRNN is again winning in recall, but losing in precision, a phenomenon that is consistently observed in both corpora and a topic worth a closer look for future work. 

Finally, in Rich ERE test data, approximately 9\% of the events are actually multi-labeled. Our current model uses softmax output layer and is thus innately incapable of making multi-label predictions. Despite this limitation, FBRNN achieved competitive result on Rich ERE with only 0.8\% difference from the best reported system in the TAC 2015 competition.

\section{Conclusions} 
This paper proposes a novel language-independent event detection method based on RNNs which can automatically extract effective features from raw text to detect event nuggets. We conducted two experiments to compare FBRNN with the state-of-the-art event detection systems on the ACE 2005 and Rich ERE 2015 corpora. These experiments demonstrate that FBRNN achieves competitive results compared to the current state-of-the-art.

\bibliography{acl2016}

\begin{thebibliography}{}

\bibitem[\protect\citename{Chen \bgroup et al.\egroup }2015]{DCNN}
Yubo Chen, Liheng Xu, Kang Liu, Daojian Zeng, and Jun Zhao.
\newblock 2015.
\newblock {Event Extraction via Dynamic Multi-Pooling Convolutional Neural
  Networks}.
\newblock {\em Association for Computational Linguistics}, 1:167--176.

\bibitem[\protect\citename{Cho \bgroup et al.\egroup }2014]{Cho}
Kyunghyun Cho, Bart van Merrienboer, {\c{C}}aglar G{\"{u}}l{\c{c}}ehre, Dzmitry
  Bahdanau, Fethi Bougares, Holger Schwenk, and Yoshua Bengio.
\newblock 2014.
\newblock {Learning Phrase Representations using {RNN} Encoder-Decoder for
  Statistical Machine Translation}.
\newblock {\em Empirical Methods in Natural Language Processing}, pages
  1724--1734.

\bibitem[\protect\citename{Collobert and Weston}2008]{Collobert_2008}
Ronan Collobert and Jason Weston.
\newblock 2008.
\newblock {A unified architecture for natural language processing: deep neural
  networks with multitask learning}.
\newblock {\em ICML}, pages 160--167.

\bibitem[\protect\citename{Cross and Huang}2016]{B_LSTM}
James Cross and Liang Huang.
\newblock 2016.
\newblock {Incremental Parsing with Minimal Features Using Bi-Directional
  LSTM}.
\newblock {\em Association for Computational Linguistics}.

\bibitem[\protect\citename{Duchi \bgroup et al.\egroup }2011]{AdaGrad}
John Duchi, Elad Hazan, and Yoram Singer.
\newblock 2011.
\newblock {Adaptive subgradient methods for online learning and stochastic
  optimization}.
\newblock {\em The Journal of Machine Learning Research}, 12:2121--2159.

\bibitem[\protect\citename{Hinton \bgroup et al.\egroup }2012]{dropout_2}
Geoffrey~E. Hinton, Nitish Srivastava, Alex Krizhevsky, Ilya Sutskever, and
  Ruslan Salakhutdinov.
\newblock 2012.
\newblock {Improving neural networks by preventing co-adaptation of feature
  detectors}.
\newblock {\em CoRR}, abs/1207.0580.

\bibitem[\protect\citename{Hochreiter and Schmidhuber}1997]{LSTM}
Sepp Hochreiter and J{\"{u}}rgen Schmidhuber.
\newblock 1997.
\newblock {Long Short-Term Memory}.
\newblock {\em Neural Computation}, 9(8):1735--1780.

\bibitem[\protect\citename{Huang and Riloff}2012]{Huang}
Ruihong Huang and Ellen Riloff.
\newblock 2012.
\newblock {Modeling Textual Cohesion for Event Extraction}.
\newblock {\em AAAI}.

\bibitem[\protect\citename{Ji and Grishman}2008]{Ji_Grashman_2008}
Heng Ji and Ralph Grishman.
\newblock 2008.
\newblock {Refining Event Extraction through Cross-Document Inference}.
\newblock {\em Association for Computational Linguistics}, pages 254--262.

\bibitem[\protect\citename{Kingma and Ba}2015]{ADAM}
Diederik Kingma and Jimmy Ba.
\newblock 2015.
\newblock {Adam: A method for stochastic optimization}.
\newblock {\em arXiv:1412.6980}.

\bibitem[\protect\citename{Li \bgroup et al.\egroup }2013a]{Li_2013_a}
Peifeng Li, Qiaoming Zhu, and Guodong Zhou.
\newblock 2013a.
\newblock {Argument Inference from Relevant Event Mentions in Chinese Argument
  Extraction}.
\newblock {\em Association for Computational Linguistics}, 1:1477--1487.

\bibitem[\protect\citename{Li \bgroup et al.\egroup }2013b]{Li_2013}
Qi~Li, Heng Ji, and Liang Huang.
\newblock 2013b.
\newblock {Joint Event Extraction via Structured Prediction with Global
  Features}.
\newblock {\em Association for Computational Linguistics}, 1:73--82.

\bibitem[\protect\citename{Li \bgroup et al.\egroup }2014]{Li_2014}
Qi~Li, Heng Ji, Yu~Hong, and Sujian Li.
\newblock 2014.
\newblock {Constructing Information Networks Using One Single Model}.
\newblock {\em Empirical Methods in Natural Language Processing}, pages
  1846--1851.

\bibitem[\protect\citename{Liao and Grishman}2010]{Liao_2010}
Shasha Liao and Ralph Grishman.
\newblock 2010.
\newblock {Using Document Level Cross-Event Inference to Improve Event
  Extraction}.
\newblock {\em Association for Computational Linguistics}, pages 789--797.

\bibitem[\protect\citename{McClosky \bgroup et al.\egroup }2011]{McClosky}
David McClosky, Mihai Surdeanu, and Christopher~D. Manning.
\newblock 2011.
\newblock {Event Extraction as Dependency Parsing}.
\newblock {\em Association for Computational Linguistics}, pages 1626--1635.

\bibitem[\protect\citename{Mikolov \bgroup et al.\egroup }2013]{word_2_vec}
Tomas Mikolov, Ilya Sutskever, Kai Chen, Gregory~S. Corrado, and Jeffrey Dean.
\newblock 2013.
\newblock {Distributed Representations of Words and Phrases and their
  Compositionality}.
\newblock {\em Neural Information Processing Systems}, pages 3111--3119.

\bibitem[\protect\citename{Mitamura \bgroup et al.\egroup }2015]{TAC_2015}
Teruko Mitamura, Zhengzhong Liu, and Eduard Hovy.
\newblock 2015.
\newblock {Overview of {TAC} {KBP} 2015 Event Nugget Track}.
\newblock {\em Text Analysis Conference}.

\bibitem[\protect\citename{Nguyen and Grishman}2015]{CNN}
Thien~Huu Nguyen and Ralph Grishman.
\newblock 2015.
\newblock {Event Detection and Domain Adaptation with Convolutional Neural
  Networks}.
\newblock {\em Association for Computational Linguistics}, 2:365–--371.

\bibitem[\protect\citename{Patwardhan and Riloff}2009]{Patwardhan}
Siddharth Patwardhan and Ellen Riloff.
\newblock 2009.
\newblock {A Unified Model of Phrasal and Sentential Evidence for Information
  Extraction}.
\newblock {\em Empirical Methods in Natural Language Processing}, pages
  151--160.

\bibitem[\protect\citename{Paulus \bgroup et al.\egroup }2014]{Paulus}
Romain Paulus, Richard Socher, and Christopher~D. Manning.
\newblock 2014.
\newblock {Global Belief Recursive Neural Networks}.
\newblock {\em Neural Information Processing Systems}, pages 2888--2896.

\bibitem[\protect\citename{Socher \bgroup et al.\egroup }2014]{Socher_2014}
Richard Socher, Andrej Karpathy, Quoc~V. Le, Christopher~D. Manning, and
  Andrew~Y. Ng.
\newblock 2014.
\newblock {Grounded Compositional Semantics for Finding and Describing Images
  with Sentences}.
\newblock {\em Transactions of the Association for Computational Linguistics},
  2:207--218.

\bibitem[\protect\citename{Srivastava \bgroup et al.\egroup }2014]{dropout}
Nitish Srivastava, Geoffrey~E. Hinton, Alex Krizhevsky, Ilya Sutskever, and
  Ruslan Salakhutdinov.
\newblock 2014.
\newblock {Dropout: a simple way to prevent neural networks from overfitting}.
\newblock {\em Journal of Machine Learning Research}, 15(1):1929--1958.

\bibitem[\protect\citename{Tai \bgroup et al.\egroup }2015]{Tai}
Kai~Sheng Tai, Richard Socher, and Christopher~D. Manning.
\newblock 2015.
\newblock {Improved Semantic Representations From Tree-Structured Long
  Short-Term Memory Networks}.
\newblock {\em Association for Computational Linguistics}, 1:1556--1566.

\bibitem[\protect\citename{Zeiler}2012]{AdaDelta}
Matthew~D Zeiler.
\newblock 2012.
\newblock {ADADELTA: an adaptive learning rate method}.
\newblock {\em arXiv preprint arXiv:1212.5701}.

\end{thebibliography}
\bibliographystyle{acl2016}

\end{document}